\documentclass[11pt,a4paper]{article}
\usepackage[hyperref]{emnlp2018}
\usepackage{times}
\usepackage{latexsym}
\usepackage{amsmath}
\usepackage{graphicx}
\usepackage{algorithmic}
\usepackage{amsmath}

\usepackage[ruled]{algorithm2e}
\usepackage[utf8]{inputenc}
\usepackage[hang,flushmargin]{footmisc}
\usepackage{lipsum}
\makeatletter
\newcommand{\appropto}{\mathrel{\vcenter{
  \offinterlineskip\halign{\hfil$##$\cr
    \propto\cr\noalign{\kern2pt}\sim\cr\noalign{\kern-2pt}}}}}

\usepackage{mathtools}

\newcommand{\algorithmfootnote}[2][\footnotesize]{%
  \let\old@algocf@finish\@algocf@finish
  \def\@algocf@finish{\old@algocf@finish
    \leavevmode\rlap{\begin{minipage}{\linewidth}
    #1#2
    \end{minipage}}%
  }%
}
\usepackage{url}
\usepackage{makecell}

\aclfinalcopy 

\title{Denoising Neural Machine Translation Training with \\ Trusted Data and Online Data Selection}

\author{
 Wei Wang \\ Google Research  \\ {\tt wangwe@google.com} \\ \And
 Taro Watanabe \\ Google Research  \\ {\tt tarow@google.com} \\ \And
 Macduff Hughes \\ Google Research  \\ {\tt macduff@google.com} \\
 \AND
 Tetsuji Nakagawa \\ Google Research  \\ {\tt tnaka@google.com} \\ \And
 Ciprian Chelba \\ Google Research  \\ {\tt ciprianchelba@google.com} \\
 }

\date{}

\begin{document}
\maketitle
\begin{abstract}
  Measuring domain relevance of data and identifying or selecting well-fit domain data for machine
  translation (MT) is a well-studied topic, but denoising is not yet. Denoising
  is concerned with a different type of data quality and
  tries to reduce the negative impact of data noise on MT training, in particular, neural MT (NMT)
  training. This paper generalizes methods for measuring and selecting data for domain MT and
  applies them to denoising NMT training.  The proposed approach uses trusted data and a
  denoising curriculum realized by online data selection.
Intrinsic and extrinsic evaluations of
  the approach show its significant effectiveness for NMT to train on data with severe noise.
\end{abstract}

\section{Introduction}\label{intro}
Data noise is an understudied topic in the machine translation (MT) field.
 Recent research has found that data noise
has a bigger impact on neural machine translation (NMT) than on statistical machine translation \cite{nmtnoise},
 but learning what data quality (or noise) means in NMT and how to make
NMT training robust to data noise remains an open research question.

On the other hand, a rich body of MT data research focuses on {\em domain data}
 relevance and selection for domain adaptation purpose. As a result,
 effective and successful methods have
been published and shown to work for both SMT and NMT. For example, \cite{Axelrod2011}
introduce a metric for measuring the data relevance to a domain by using n-gram language
models (LM).  \cite{dynamiccds} employ a neural-network version of it and propose
a gradually-refining strategy to dynamically schedule data during NMT training. In these methods,
a large amount of in-domain data are used to help measure data domain relevance.

Data noise is a different quality that has been shown to affect NMT performance in particular.
 In MT, the use of web crawl, automatic methods for parallel data mining, sentence alignment
provide us with parallel data of variable quality from many points of view: sentence breaking,
 poor sentence alignments, translations, domain adequacy, tokenization and so forth.
 To deal with such data noise, a commonly used
 practice is (static) data filtering with simple heuristics or classification.
The NMT community increasingly realizes that this type of quality matters for general NMT
translation accuracy.  For example, \cite{nmtnoise} studies the types of data noise and their impact on
NMT;
WMT 2018 introduces a Parallel Corpus Filtering task
on noisy web-crawled data.

Unfortunately, the ingredients that made domain data selection methods successful
have not been studied in the NMT denoising context. Specifically,
\begin{itemize}
\item How to measure noise?
\item How does noise dynamically interact with the training progress?
\item How to denoise the model training with a small, trusted parallel dataset?
\end{itemize}
In the denoising scenario, the trusted data would be the counterpart of in-domain
monolingual data of domain data selection. Trusted data can be human translations,
a small amount of which can be easily available as a development set or validation set
from a normal MT setup.

We use the example in Table~\ref{noisy_example} to illustrate
 the challenges in the NMT denoising problem, as well as
 the issue of directly applying existing domain methods as is for this purpose.
Both sentences in the example appear to be relevant to travel conversations, but
the sentence pair is ``noisy'' in that, a part of the English sentence 
does not align to anything on the Chinese side, yet the pair contains some translation and the
sentences are fluent.
An LM-based domain-data selection method would generally treat it as a suitable domain example
for building a travel NMT model and may not consider this noise.

\begin{table}[t]
\centering
  \begin{tabular}{ll}
    \hline
    zh & gongche zhan zai nali? \\
    {\it zh-gloss} & {\it bus \hspace{.25in} stop \hspace{.03in} is \hspace{0.04in} where?} \\
    en & Where is the bus stop? For bus 81. \\
    \hline
  \end{tabular}
\caption{A noisy sentence pair. \label{noisy_example}}
\end{table}

A simple data filtering method based on length or a bilingual dictionary
can easily filter it, but, intuitively, the example may still be
useful for training the NMT model, especially in a data-scarce scenario --
the Chinese sentence and the first half of the English sentence are still a translation pair.
This suggests the subtlety in identifying noisy data for MT -- It is not a simple binary
problem: Some training samples may be partially useful to training a model, and their usefulness
 may also change as training progresses.

An NMT model alone may be incapable of identifying noise. Under a conditional seq2seq
NMT model that translates Chinese into English, a word, e.g., 81, in the
extra English fragment may receive a low probability (or a high loss), but 
that could as well mean that is hard but still correct translation. Here
is then where the trusted data can play a role -- It can help produce a (slightly)
 better model for the first model to compare against to be able to distinguish
 informative hard examples from harmful noisy ones.

In this paper, we propose an approach to denoising online NMT training.  It uses
a small amount of trusted data to help models measure noise in a sentence pair. The noise
is defined based on comparison between a pair of a noisy NMT model and another, slightly
denoised NMT model, inspired by the contrastive in-domain LM vs out-of-domain LM idea.
It employs online data selection to sort sentence pairs by noise level so that the 
model is trained on gradually noise-reduced data batches.
We show that language model based domain data selection method
 as is does not work well whereas the proposed approach
is quite effective in denoising NMT training.


\section{Related Research}\label{related}
One line of research that is related to our work is data selection for machine translation.
It has been mostly studied in the domain adaptation context. Under this context,
a popular metric to measure domain relevance of data is based on cross entropy difference (CED)
between an in-domain and an out-of-domain language models.  For example, \citep{Moore2010}
 selects LM training data with CED
 according to an in-domain LM and a generic LM.
\citep{Axelrod2011} propose the contrastive data selection idea to select parallel domain data.
It ranks data by the bilingual CED that is computed, for each language,
 with a generic n-gram LM and a domain one. Even more recently, \cite{dynamiccds} employ
a neural-network version  of it along with a dynamic data selection idea and achieve better
domain data selection outcome. \citep{LM_M1} compute the CED using IBM translation Model 1
and achieve the best domain data selection/filtering effect for SMT combined with LM selection;
 The case of partial or misalignments with a bilingual scoring mechanism rather than LMs is also discussed.

Another effective method to distinguish domain relevance is to build a classifier.
A small amount of trusted parallel data is used in classifier training.
 For example,
\cite{chen2016a} use semi-supervised convolutional neural networks (CNNs) as LMs to select domain data.
Trusted data is used to adapt the classifier/selector.
\citep{chen2016b} introduce a bilingual data selection method that uses CNNs on bitokens; The method uses
parallel trusted data and is targeted at selecting data to improve SMT; In addition to domain relevance, the work
also examines its noise-screening capability;  The method is tried on NMT and does not seem to improve.

Previous work on domain data selection has shown that the order in which data are scheduled matters a lot for
NMT training, a research that is relevant to curriculum learning \cite{curri_learn} in machine
learning literature. \cite{dynamiccds} show the effectiveness of a nice ``gradually-refining'' dynamic data
 schedule.  \citep{model_stacking} find the usefulness of a similar idea,
 called model stacking for NMT domain adaptation.
Data ordering could be viewed as a way of data weighting, which can be also done by example weighting/mixing, e.g.,
\cite{D17-1155,mixing,matsoukas-rosti-zhang:2009:EMNLP}. In the context of denoising, the quality that the ordering
uses would be the amount of noise in a sentence pair, not (only)
 how much the data fits the domain of interest.

SMT models tend to be fairly robust to data noise and denoising in SMT
seems to have been a lightly studied topic.  For example,
\citep{Mediani17} uses a small, clean seed corpus and designs classifier filter
 to identify noisy data with lexical features; and also there is a nice list of
works accumulated over years, compiled on the SMT Research Survey
 Wiki\footnote{\url{http://www.statmt.org/survey/Topic/CorpusCleaning}}.

The importance of NMT denoising has been increasingly realized.
\cite{nmtnoise} study the impact of five types of artificial noise in parallel data
on NMT training and find that NMT is less robust to data noise. \cite{N18-1136}
select well-translated examples by identifying semantic divergences in parallel data.
\cite{DBLP:journals/corr/abs-1711-00043} bootstrap backtranslations with a denoising loss term, in
an unsupervised NMT context.  Label noise is also a generally studied topic, e.g., \cite{NIPS2013_5073}.

In a sense, our approach is an application of active learning \citep{Settles10activelearning}. Active learning
is usually employed for the model to interactively choose novel examples to obtain labels for further training
a given model.
 In our case we use the idea to select the already labeled data that the model finds
 useful at a given point during training. The usefulness signal is guided by a small trusted dataset.







\section{Online NMT Training} \label{sec:ds_as_search}
We usually train NMT models with online optimization,  e.g., stochastic gradient descent.
 At a time step $t$, we have an
NMT model $p(y|x; \theta_t)$ translating from sentence $x$ to $y$ with parameterization
$\theta_t$.
The model choice could be, for example, RNN-based \citep{wu2016}, CNN-based \citep{DBLP:journals/corr/GehringAGYD17},
Transformer model \citep{Vaswani2017} or RNMT+ \citep{Chen2018}.
To move $p(y|x; \theta_t)$ to next step, $t+1$, a {\em random} data batch $b_t$ is normally used
to compute the cross entropy loss.
  The prediction accuracy of $p(y|x; \theta_t)$ does not depend on the data of this batch alone,
but on all data the model has seen so far.

\section{The Denoising Problem} \label{problem}
The problem we address in the paper is as follows.
We have a large, noisy, mixed-domain dataset $\widetilde{D}$ whose size
is on the order of hundreds of millions of sentence pairs or larger.
An NMT model trained on this noisy data may suffer from low translation accuracy or severe translation errors.
We also have a small trusted dataset $\widehat{D}$ consisting of several thousand sentence pairs or even less.
 We address the denoising scenario where the trust fraction $|\widehat{D}| / |\widetilde{D}| \ll 1$
 ($|\widehat{D}|$ being the size of $\widehat{D}$).

Trusted data can be human translations or any other source of parallel data of higher quality than
 the translations produced by our model.  The trusted data we use in experiments contains noise, too.
 We think that, for the trusted data to improve, it needs to be
stronger than the translation quality from the model we are improving, and as we
will show, we define the noise level of a sentence pair relative to a model.

 We are concerned with a method for selecting noise-reduced data batches to train the NMT model using
online training.
The trusted data is used to help measure data noise in a sentence pair. Training data is digested by
training in terms of (cross entropy) loss, thus selecting noise-reduced sentence pairs to train on would be
equivalent to denoising the training loss term (thus the training process).

\section{Our Approach} \label{approach}
We first define how to measure noise with the help of the small trusted dataset. Then we use
 it to control the schedule of the data batches to train the NMT model. 

\subsection{Incremental denoising with trusted data} \label{sec:fine-tuning}
Given a model $p(y|x; \widetilde{\theta})$ trained on noisy data $\widetilde{D}$,
a practical way to denoise it with a small amount of
trusted data $\widehat{D}$ would be to simply fine-tune the model on the trusted data,
considering that a small trusted dataset alone is not enough to reliably train an NMT model
from scratch.
Fine-tuning has been used in NMT domain adaptation to adapt an existing
NMT model on a small amount of in-domain data, for example, in \cite{dynamiccds}.
We hypothesize that it would be effective for denoising, too, which will be verified
by our experiments.

To facilitate the introduction of our denoising method, we introduce a {\em denoise}
function that denoises a model, $p(y|x; \widetilde{\theta})$, on the trusted data $\widehat{D}$
by fune-tuning:

\begin{eqnarray}
  p(y|x; \widehat{\theta}) &=& \textrm{denoise}\left( p(y|x; \widetilde{\theta}), \widehat{D}\right) \label{denoise}
\end{eqnarray}
Eq~\ref{denoise} represents that model $p(y|x; \widetilde{\theta})$ with initial parameterization $\widetilde{\theta}$ is
fine-tuned on the trusted data $\widehat{D}$ to yield a
denoised model, $p(y|x; \widehat{\theta})$. With a small trusted dataset,
the fine-tuning may take a small number of training steps.

\subsection{Definition of data noise} \label{sec:incr}
MT training samples can be noisy in many ways, and different types of noise
might have different impact on NMT. Furthermore, human's definition of data noise may
not be completely consistent with NMT model's perspective. Therefore, instead of defining noise
in these aspects, we 
 simply use model probabilities and rely on the quality of
the trusted data.
 After all, data needs to be ingested by model training via (cross-entropy) loss.

Supposed we are given a {\em noisy model}, $p(y|x, \widetilde{\theta})$, that has been trained on
noisy data and a {\em denoised model}, $p(y|x, \widehat{\theta})$, obtained by Eq~\ref{denoise},
with the denoised model being a slightly more accurate probability distribution than the noisy version.
For a sentence pair $(x, y)$ of a source sentence $x$ and its target translation $y$,
 we can compute its ``noisy logprob'' under the noisy model:
\begin{eqnarray}
  L_{p(y|x, \widetilde{\theta})} &=& \log p(y|x, \widetilde{\theta}) \label{noisy_loss}
\end{eqnarray}
We can also compute its ``denoised logprob'' under the denoised model:
\begin{eqnarray}
  L_{p(y|x, \widehat{\theta})} &=& \log p(y|x, \widehat{\theta}) \label{denoised_loss}
\end{eqnarray}
We then define the noise level of a sentence pair $(x, y)$ as the difference of a noisy model score over
a denoised model score:
\begin{eqnarray}
  \textrm{noise}(x, y; \widetilde{\theta}, \widehat{\theta}) & =& L_{p(y|x; \widetilde{\theta}_t)} - L_{p(y|x; \widehat{\theta}_t)} \label{noise}
\end{eqnarray}
The noise level of a sentence pair is the sum of the per-word noise over all the target words (under conditional
translation models).
$\textrm{Noise}(x, y; \widetilde{\theta}, \widehat{\theta})$ could also be normalized by the length of
sentence $y$ empirically.
The bigger $\textrm{noise}(x, y; \widetilde{\theta}, \widehat{\theta})$ is, the higher noise level the
sentence pair has. A negative value of $\textrm{noise}(x, y; \widetilde{\theta}, \widehat{\theta})$
 means that the sentence pair has more information to
learn from (cleaner).



The noise in a sentence pair is defined in terms of the comparison between two models: the noisy model,
$\widetilde{\theta}$, and the denoised model, $\widehat{\theta}$. Under this definition, noise is relative -- 
A sentence pair could have negative $\textrm{noise}(x, y; \widetilde{\theta}, \widehat{\theta})$ (not noise) for weeker models (i.e., an earlier checkpoint of $\widetilde{\theta}$ in an NMT training), but could become noisy (positive value) for stronger models (i.e., a later checkpoint of $\widetilde{\theta}$). This would address
one of the issues we illustrated in Section~\ref{intro} with the example in Table~\ref{noisy_example}.

Notice that this definition of noise is a generalization of the bilingual cross-entropy difference (CED)
defined and used in \cite{Axelrod2011,dynamiccds} to measure domain relevance of a sentence pair.
 We use seq2seq NMT models to directly model
a sentence pair, while previous works use language models to model monolingual sentences independently.
A language model corresponds just to the decoder component of a translation model and thus cannot model
the translation quality. The lack of the encoder component (thus translation) makes the
 LM-based method unsuitable for denoising, as we show in experiments.
 Additionally, we use
a small, bilingual trusted dataset (semi-supervision) rather than lots of in-domain data (heavier
supervision).

\subsection{Denoising by online data selection} \label{sec:contrastive_training}
\subsubsection{The idea}\label{idea}
Our main idea for online denoising of NMT training is to train an NMT model
on a progressively-denoised curriculum (data batches). As a result,
 the entire training becomes a continuous fine-tuning.
 We realize the denoising curriculum through
 dynamic data selection to ``anneal'' the noise level in data batches
 over training steps. Therefore,
our method tries to control the way how
 noise dynamically interacts with training loss by data selection,
instead of directly altering per-example loss. The assumption
is that $\widetilde{D}$ contains good examples to select, which
is usally true with a big enough training dataset $\widetilde{D}$.

More concretely,  at each step with an initial (potentially still noisy)
 model, $p(y|x; \widetilde{\theta}_t)$, the method denoises it
(by Eq~\ref{denoise}) with the trusted data $\widehat{D}$ into a slightly better model
 $p(y|x; \widehat{\theta}_t)$ for that step. With this pair of noisy and denoised models,
we then compute noise scores for examples in  a
buffer $\widetilde{B}_{t}^{\textrm{random}}$ that is randomly drawn from $\widetilde{D}$
per step
 and maintained during training. We sort the noise scores.
 The final, actual data
batch $b_t$ is then randomly sampled from the top $r_t$ portion of $B_t^{\textrm{random}}$ based
on the sorted scores, where $r_t$, called {\em selection ratio}, is increasingly tightened.
Averaged noise level of examples in the top $r_t$ portion expects to become less over time. As
a result, the data batches $b_t$'s that are actually fed to train the final model are
gradually denoised. Algorithm~\ref{active_CDS2} summarizes the idea.
It is worth pointing out that this denoising method is realized by
 a bootstrapping process, in which, $\widetilde{\theta}_t$
and $\widehat{\theta}_t$ iteratively bootstrap each other
 by interacting with the trusted data and selected denoised data.



\begin{algorithm}[t]
  \label{active_CDS2}
  \begin{algorithmic}[1]
    \STATE {\bf Input}: Noisy data $\widetilde{D}$, trusted data $\widehat{D}$
    \STATE {\bf Output}: A denoised, better model
    \STATE $t=0$; Randomly initialize $\widetilde{\theta}_0$.
    \WHILE{ $t < T$}
      \STATE  $p(y|x; \widehat{\theta}_t)$ $\leftarrow$
          \textrm{denoise}($p(y|x; \widetilde{\theta}_t)$, $\widehat{D}$).
      \STATE Randomly draw $\widetilde{B}_t^{\textrm{random}}$ from $\widetilde{D}$.
      \STATE Compute noise for examples in $\widetilde{B}_t^{\textrm{random}}$ by Eq~\ref{noise}.
      \STATE Sort $\widetilde{B}_t^{\textrm{random}}$ by noise scores.
      \STATE Sample $b_t$ from top $r_t$ of above sorted buffer.
      \STATE Train $p(y|x; \widetilde{\theta}_t)$ on $b_t$ to produce new model $p(y|x; \widetilde{\theta}_{t+1})$.
      \STATE Discard the denoised model $p(y|x; \widehat{\theta}_t)$.
      \STATE $t \leftarrow$ $t+1$.
    \ENDWHILE
  \end{algorithmic}
  \caption{Denoising NMT training with trusted data and online data selection. \label{active_CDS2}}
  \label{active_CDS2}
\end{algorithm}

We choose to use the following exponential decaying function
for selection ratio, $r_t$, to anneal data noise by data selection\footnote{
We simply use one of the ways to anneal learning rate as the decaying function to anneal
 training data selection.
}:
\begin{eqnarray}
  r_t  = 0.5^{t/H} \label{ratio}
\end{eqnarray}
It keeps decreasing/tightening over time $t$.  The entire training
 thus becomes a continuous fine-tuning process,
 in a self-paced learning \cite{NIPS2010_3923} fashion.

In Equation~\ref{ratio}, $H$ is a
 hyper-parameter controlling the decaying pace: It halves $r_t$ every $H$ steps.
For instance, 
$H=10^6$ means that, at step 1 million, data batch $b_t$ will be drawn from the top-50\% out of
sorted buffer.

In practice, it may be desirable to set a floor value for $r_t$ (e.g., 0.2) to avoid potential selection bias.
$B_t^{\textrm{random}}$ needs also to be big enough such that there are enough examples in the top $r_t$
range to select from to form the final training batch $b_t$, which is usally a constant size --
It needs to contain at least $|b_t|/ r_{\textrm{floor}}$ examples.  

The noise annealing is inspired by \cite{dynamiccds},
 but we anneal data quality at per step to make the approach more friendly to 
 NMT online optimization, instead of per data epoch.
 Compared to static selection, the noise annealing idea also makes every
 training example useful, by digesting noisy examples earlier and fine-tuning on 
good-quality examples later on.



Note that there are two reasons that this process does not overfit on the trusted data, even though it is kept being used to denoise the initial model at every step. First, the noisy model, $p(y|x; \widetilde{\theta})$ being trained over steps is never trained on the trusted data -- It is the denoised model, $p(y|x; \widehat{\theta})$,
that is trained on it and then gets discarded at the end of that step.
 Second, the online data selection progressively anneals from
 noisy examples to less noisy ones, instead of greedily keeping selecting out of the least noisy examples.

\subsubsection{Data selection per-step overhead}\label{overhead}
Compared to normal NMT training, there is a per-step data selection overhead in
Algorithm~\ref{active_CDS2}. The overhead includes (1) training the denoised model on a small trusted
dataset, which requires a small number of training steps; and (2) scoring all examples in the random buffer $B_t^{\textrm{random}}$
 with both the noisy model and the denoised model.  Both cases will in general depend on model size,
 but will probably depend even more on model type and configuration.

\subsubsection{Lightweight implementation}\label{practical}
We make Algorithm~\ref{active_CDS2} more lightweight by
decoupling model training from example noise scoring: We score all examples in $\widetilde{D}$ offline
and use scores for online data selection.

 Algorithm~\ref{active_CDS} shows the details of this idea.
 To enable offline scoring, we train the noisy model
and the denoised model prior to the final, target training, on the noisy data $\widetilde{D}$
 and the trusted data $\widehat{D}$, respectively.
We then use this pair of models to score all examples in $\widetilde{D}$ and save the scores.
In target model training, the example are retrieved into the buffer with scores, without the need
of computing on the fly. Then the remaining is similar to Algorithm~\ref{active_CDS2}.
 This effectively turns the per-step data selection overhead in Algorithm~\ref{active_CDS2}
into constant overhead.

 We can also use smaller parameterization for the noisy model and denoised model than the target model.
This may not affect their noise-discerning capability as long as they are still seq2seq models,
 the same as the target model. This is because we define the noise score in terms of logprob difference 
 and use the scores for ranking/selection (e.g., via top $r_t$),

In summary, here is the lightweight method that we eventually use to denoise NMT training with trusted data and online data selection:
Train $p(y|x; \widetilde{\theta})$ on noisy data $\widetilde{D}$ with a small parametrization.
Denoise $p(y|x; \widetilde{\theta})$ on trusted data $\widehat{D}$ to produce denoised model
 $p(y|x; \widehat{\theta})$ (Eq~\ref{denoise}).
Score the entire noisy data $\widetilde{D}$ with the above two models by Eq~\ref{noise}.
Train the target model with the above online, dynamic data selection.
Algorithm~\ref{active_CDS} describes the idea.

\begin{algorithm}[t]
  \label{active_CDS}
  \begin{algorithmic}[1]
    \STATE {\bf Input}: Noisy data $\widetilde{D}$, trusted data $\widehat{D}$
    \STATE {\bf Output}: A denoised, better model with learned parameters $\Theta$.
     
    \STATE Train $p(y|x; \widetilde{\theta})$ with small $\widetilde{\theta}$ on $\widetilde{D}$.
    \STATE $p(y|x; \widehat{\theta})\leftarrow \textrm{denoise}\left(p(y|x; \widetilde{\theta}); \hat{D}\right)$.
    \STATE Score $\widetilde{D}$ with $\widetilde{\theta}$ and $\widehat{\theta}$  by  Eq~\ref{noise}.
    \STATE $t=0$; Randomly initialize $\widetilde{\Theta}_0$.

    \WHILE{ $t < T$}
      \STATE Randomly sample $\widetilde{B}_t^{\textrm{random}}$ from $\widetilde{D}$.
      \STATE Sort $\widetilde{B}_t^{\textrm{random}}$ by offline-computed noise scores.
      \STATE Sample $b_t$ from top $r_t$ of above sorted buffer.
      \STATE Train $p(y|x; \widetilde{\Theta}_t)$ on $b_t$ to produce new model $p(y|x; \widetilde{\Theta}_{t+1})$.
      \STATE $t \leftarrow$ $t+1$.
    \ENDWHILE
  \end{algorithmic}
  \caption{Lightweight implementation of Algorithm~\ref{active_CDS2}. Actually used in experiments.
 \label{active_CDS}}
  \label{active_CDS}
\end{algorithm}

We are going to use this implementation in experiments. Note, however, that we find that 
the general method in Algorithm~\ref{active_CDS2} is very useful in understanding the nature of the
denoising problem and thus cannot be ignored in the context. 
For example, it makes us realize the denoising problem is about how to (actively) meet what
the model needs, i.e., not standalone filtering.  And also,
the bootstrapping behavior in Algorithm~\ref{active_CDS2} further motivates
 the use of the noise-annealing online data selection strategy
 and helps refine the lightweight implementation.

\section{Experiments}\label{exp}
\subsection{Setup} \label{sec:setup}

We carry out experiments for en/fr with two training datasets ($\widetilde{D}$), respectively.
Paracrawl\footnote{\url{ http://statmt.org/paracrawl}}
en/fr training raw data has 4 billion sentence pairs. After removing identities and empty source/target, about 300
million (M) sentence pairs are left.
WMT 2014 en/fr training data has about 36M sentence pairs, with provided sentence alignment. 


WMT newstest 2012-2013 is used as the development set for early stopping of training.
We use three test sets: WMT (n)ewstest 2014 ({\bf n2014}),
news (d)iscussion test 2015 ({\bf d2015}),
and a 2000-line patent test set ({\bf patent})\footnote{Obtained from \url{ https://www.epo.org.}}.
More test sets than just n2014 are used in order to confirm that
the gain obtained is not only from news domain adaptation but cross-domain, general accuracy improvement.

The WMT newstest 2010-2011 is used as the trusted data. It contains 5500 sentence pairs.
We acknowledge that ideal trusted data would probably be both well-translated and domain-matched,
but we leave the study of trusted data properties to future research.

We compute the {\em detokenized} and mixed-cased BLEU scores against the original references (per \citep{clarifybleu})
with an in-house implementation of script {\tt mteval-v14.pl}.


We use an RNN-based NMT architecture similar to \citep{wu2016}.
Our final model has 8 layers of encoder/decoder, 1024 dimensions with 512-dimension attention.
 The smaller selector (noisy and denoised) models (in Algorithm~\ref{active_CDS}) are
of 3 layers and 512 dimensions.\footnote{
Even smaller models like 2-layer x 256-dimension works, too, when we examined on an
internal dataset.
}

Denoising a model on the small trusted dataset is done by fine-tuning on it
by SGD. The training is terminated with early stopping by checking the perplexity on
the development set. It is a tiny dataset, but as we will show, its denoising impact is
quite impressive and surprising. Training on such a small data can easily overfit, we thus use a very
small learning rate 5e-5 so that the training progresses slow enough for us to reliably
 catch a good checkpoint before training stops.

In Paracrawl trainings, we train for 3M steps using SGD with learning rate 0.5 and start to anneal/reduce
the learning rate at step 2M by halving it every 200k steps.
In WMT training, we train for 2M steps with learning rate 0.5 but start to anneal learning 
rate at step 1.2M with the same pace. We use dropout 0.2 for
the WMT training. We did not use dropout for Paracrawl training due to its large training data amount.

To dynamically anneal the data batch quality (Eq~\ref{ratio}),  we set hyper-parameter $H$ to step 1.1M.
 0.2 is used as the floor selection ratio, $r_t$.  The rationale for the choice of $H$ is so that
 when learning rate annealing happens, $r_t$ is close to its minimum value to ensure the training
 is indeed trained on the desired, best selected data.

\subsection{Training data cleanness}
To measure how noisy the datasets are, we randomly sample
2000 sentence pairs from the WMT dataset. Human raters were asked to label each sentence pair 
with scales in Table~\ref{scales}.
\begin{table*}[t]
\centering
  \begin{tabular}{ll}
    \hline
    Rating scale & Explanation \\
    \hline
    \hline
    4 (Perfect) &  Almost all information (90-100\%) in the sentences is conveyed in each other. \\
    3 (Good) & Most information (70-90\%) in the sentences is conveyed. \\
    2 (Not good)& Some (30-70\%) information in the sentences is conveyed, but some is not. \\
    1 (Bad) & (10-30\%) A large amount of information in the sentences is lost or misinterpreted. \\
    0 (Poor) & (0-10\%) The two sentences are nearly or completely unrelated, or in wrong languages. \\
    \hline
  \end{tabular}
\caption{Scales for human rating sentence pairs. Percentage ranges refer to
the amount of words well translated across sentences in a pair. \label{scales}}
\end{table*}

These ratings generally reflect how well-translated a sentence pair is, however, a rating 4 does
not necessarily mean that is exactly the type of data a model needs -- Model's perspective
on good data may not completely consistent with human, because these ratings are
not necessarily connected to data loss of a model. We use these ratings mainly to assess if our
noise definition correlates to these ratings to some extent, but the noise definition could
do more. The rater agreement on good ($>= 3$) or bad ($< 3$) is $70\%$ and we find the averaged rating
is very reliable and stable to measure a small sentence pair sample.

Table~\ref{cleanness_ratings} shows that WMT 2014 data is relatively clean: it has 40\% rated as perfect; its averaged rating is 3.0 (4 being perfect).
Noise introduced by sentence alignment accounts for part of the low ratings.
We did not rate a Paracrawl sample, since just eyeballing a sample of the data reveals that it was
 noisy consisting of many boilerplates, wrong language identification, wrong translations.

\begin{table}[t]
\centering
  \begin{tabular}{ccccccc}
    \hline
    Rating scale & 4& 3 &  2 & 1 & 0 \\
    \hline
    \hline
    WMT & 47\% & 31\%  & 10\% & 3\% & 9\% \\
    \hline
  \end{tabular}
\caption{Rating stats on an en/fr WMT training data sample of 2000 sentence pairs. \label{cleanness_ratings}}
\end{table}


\subsection{Noise score vs human rating}
We expect the noise definition (Eq~\ref{noise}) to correlate with the averaged cleanness of selected data
and the dynamic scheduling method schedules data from noisy to clean. We verify this on the
sample with human ratings.

We carry out steps 1 and 2 of the practical implementation in Section~\ref{practical} to produce
the small noisy model and its denoised model. Recall that they are used to compute the noise
in each sentence pair by Eq~\ref{noise}. We repeat this on the Paracrawl data and the WMT data, respectively, and
thus we have two pairs of models, one for each dataset.

\begin{figure}[t]
\includegraphics[width=8cm,height=7cm]{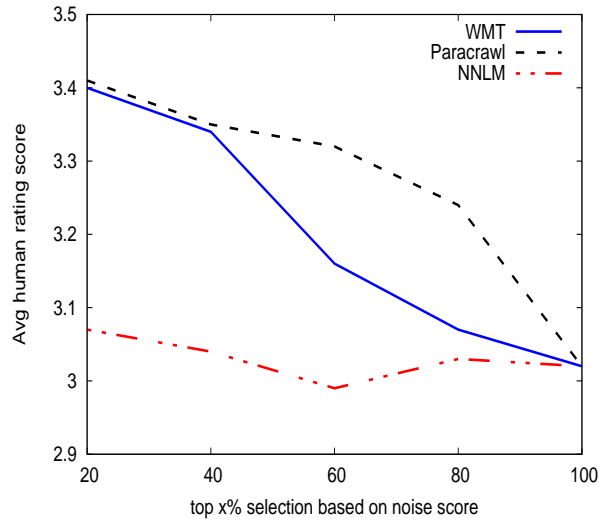}
  \caption{Noise-discerning capability of different noise scoring models.
   Curves are drawn by selecting, according to Eq~\ref{noise}, top $x\%$ (x-axis) out of a rated sample of 2000 random
sentence pairs from the WMT en/fr dataset. WMT: noise scoring models trained on WMT training data, and trusted data. Paracrawl: noise scoring models trained on Paracrawl data, and trusted data. NNLM: neural net based LM selection models trained on Paracrawl data, and trusted data. Trusted data are the same dataset. \label{ranking}
}
\end{figure}
We apply each pair of models to score the rated WMT sample, sort the sentence pairs by noise scores.
We then select $x\%$ least noisy sentence pairs. Each $x\%$ corresponds to a subset and we
compute the averaged human rating for that subset.
In Figure~\ref{ranking}, x-axis shows $x\%$, the percentage out of the entire sample; y-axis
shows the averaged human rating for the $x\%$ selection subset.
Going from right to left, data indeed becomes cleaner as selection becomes tighter for the scoring
models in our proposed method: WMT is noise scoring models trained on WMT
training data, and trusted data. Paracrawl is the noise scoring models trained on Paracrawl data,
 and trusted data. Trusted data are the same dataset. We explain the dot-dashed line in a
 later experiment (Section~\ref{sec:nnlm}).

Ranking capability of the Paracrawl selector seems slightly
better than the WMT one in discerning noisier sentence pairs. We speculate this is because the noisy Paracrawl data
``amplifies'' the contrastive effect of the pair of models.

\subsection{BLEU scores}
\begin{table}[t]
\centering
  \begin{tabular}{lccc}
    \hline
    System & n2014 & d2015& patent\\
    \hline
    \hline
    Paracrawl dataset && \\
    P1 Paracrawl baseline & 31.6 & 30.7 &  37.2 \\
    P2 Incr-denoise P1 & 34.0 & 33.7   & 44.7 \\
    P3 Online denoised    & {\bf 35.2}  & {\bf 35.6}  & {\bf 46.9} \\
    \hline
    WMT dataset && \\
    W1 WMT baseline &  36.2 & 35.8 & 45.7 \\
    W2 Incr-denoise W1 & 36.2 & 35.8 & 45.7 \\
    W3 Online denoised & {\bf 36.9} & {\bf 36.4} & {\bf 46.1} \\
    \hline
  \end{tabular}
  \caption{BLEU scores of Denoising experiments with en/fr Paracrawl data and WMT data. 
           ``Incr-denoise P1'' refers to applying the incremental denoising on
            the noisy baseline P1 with method in Section~\ref{sec:fine-tuning}. Similarly for ``incr-denoise W1''.
           Under paired bootstrapped test at $p<0.05$, P3 is significantly better than P2, P3 than P1,
            P2 than P1, on all test sets. W3 is significantly better than W1 on n2014.
   \label{paracrawl_bleus}}
\end{table}

BLEU scores in Table~\ref{paracrawl_bleus} show the impact of denoising. For each training dataset, we
have three experiments: baseline, noisy training with random data batch selection (P1 and W1); Denoising baseline with trusted data by fine-tuning
the baseline on it (Eq~\ref{denoise}) (P2 and W2); Training a model from scratch with online 
 training by dynamic, gradually noise-reduced data selection (P3 and W3).

First of all, P1 vs P2, it is impressive that just fine-tuning a noisy baseline on a small trusted dataset
yiels a big impact. P2 improves P1 by +2.4 BLEU on n2014, +3 BLEU on d2015 and +7.5 BLEU on patent.
The Paracrawl experiments and the above rating ranking curves (Figure~\ref{ranking})
 indicate the power of simple incremental denoising on trusted data (Section~\ref{sec:fine-tuning}) when
the background data is very noisy.
 In NMT domain adaptation literature (e.g., \cite{dynamiccds}), it is known that
fine-tuning on domain data improves domain test sets, but it is also known that it may hurt test sets that are
 out of domain (forgetting). We think our experiments are the first to report the incremental denoising power of
fine-tuning on a tiny trusted data. Notice incremental denoising does not
improve on WMT data (W1 vs W2) probably because WMT data is relatively cleaner. This, however, would
indicate that the gain for P1 vs P2 is less likely a domain adaptation effect.

P2 vs P3 shows that the online denoising approach reduced the training noise further more and
 gains +1.2 n2014 BLEU, +1.9 d2015 BLEU and +2.2 patent BLEU, on top of incremental denoising on trusted data.
On the WMT dataset, W2 vs W3 shows that, even though the trusted data does not directly help, the online
denoising helps by +0.7 n2014 BLEU, +0.6 d2015 BLEU and +0.4 patent BLEU. 
We carried out paired bootstrapped statistical significance test \cite{koehn:2004:EMNLP} between systems, at $p<0.05$,
P3 is significantly better than P2, P3 than P1, P2 than P1, across all test sets; W3 is significantly better
than W1 only on n2014.

We also would like to note the strength of the WMT baseline system (W1).
Its n2014 BLEU is 36.2, detokenized, case-sensitive.
Published literatures tend to report tokenized, case-sensitive BLEU scores, for
which W1 BLEU becomes 40.2 on the same test set. This is a strong score
with a standard LSTM RNN network, compared to published results for this task.

\subsection{Data order}
Our online denoising method dynamically selects data batches whose noise
is gradually reduced to train the target model.  We carry out two sets of
 experiments to prove that this is necessary for denoising. 

In the first experiment, we compare P3 (in proposed data order) to the ``reverse'' of P3, where data batches
are dynamically scheduled in a reverse, noise gradually increasing order such that
the model is trained on cleaner data earlier and then noisier data later (i.e., by simply
 flipping the sign of Eq~\ref{noise}) -- The entire training then becomes a continuous reverse
fine-tuning. Table~\ref{order} shows that the reverse order (P4) clearly does not work as
 effective for denoising, even though P4 still slightly improves the baseline
with random data selection (P1 in Table~\ref{paracrawl_bleus}).

\begin{table}[t]
\centering
  \begin{tabular}{lccc}
    \hline
    System & n2014 & d2015& patent\\
    \hline
    \hline
    Paracrawl dataset && \\
    P1 Random order & 31.6 & 30.7 &  37.2 \\
    P3 Online denoised & {\bf 35.2}  & {\bf 35.6}  & {\bf 46.9} \\
    P4 Reverse order of P3 & 32.6 & 31.1 & 40.9 \\
    \hline
  \end{tabular}
  \caption{Online denoising: NMT trained on data sorted according to noisiness level.
 P3 is trained on noisier to cleaner data order. Reversely, P4 is trained on cleaner to
noisier data order.  \label{order}}.
\end{table}

In another experiment, we select 3 data subsets based on the amount of noise
in each sentence pair, each subset being noise-reduced to different degree.
For example, we select top 80\% least noisy sentence pairs (denoted as $S_{80\%}$)
out the entire Paracrawl data. Then we select the top half of $S_{80\%}$ which
is essentially $40\%$ of the Paracrawl data. We denote it as $S_{40\%}$,
 similarly, $S_{20\%}$, therefore 
$S_{80\%} \supset S_{40\%} \supset S_{20\%}$.
And we expect the averaged noise in the smaller percentage would be less according
to Figure~\ref{ranking}.
Then we fine-tune P1 (noisy baseline) on $S_{80\%}$ with early stopping on devset, followed by
the fine-tuning on $S_{40\%}$ and then $S_{20\%}$. Table~\ref{rocket_fuel} shows
that each stricter subset is able to boost the previous training across all test sets,
 by further denoising. This also confirms the importance of the right data order in denoising.
\begin{table}[t]
\centering
  \begin{tabular}{llcccc}
    \hline
    &Subset& n2014 & d2015& patent \\
    \hline
    \hline
    P1& & 31.6 & 30.7 &  37.2 \\
    P5& $S_{80\%}$ & 33.1 & 32.3   & 44.3 \\
    P6& $S_{40\%}$ & 33.9 & 34.4   & 45.1 \\
    P7& $S_{20\%}$ & 34.4 & 34.6   & 45.6 \\
    \hline
  \end{tabular}
\caption{Nested datasets: Data order is important for denoising.
$S_{80\%} \supset S_{40\%} \supset S_{20\%}$ with stricter/smaller set less noisy.   
  \label{rocket_fuel}}
\end{table}

P3 vs P4 seems to confirm the spirit of Curriculum Learning \cite{curri_learn} --
CL promotes ordering data to gradually focus on those most important examples,
and here the training has a better outcome (P3) by training on progressively
noised-reduced data.

\subsection{Language model selection}\label{sec:nnlm}
The proposed method uses seq2seq NMT models for online data selection.  We can
replace them with neural network language models (NNLM) with everything
else the same, to confirm that the LM based method that is
popular for domain data selection is not designed for denoising.

We first check if the NNLM selection scores correlate with human ratings.
 As shown by the dot-dashed line (red) in Figure~\ref{ranking}, it does not seem to
 -- As we tighten the selection percentage (from right to left),
the averaged rating of sentence pairs falling into that percentage does not 
increase,  but the method that employs the seq2seq models to compute noise
scores (Eq~\ref{noise}) does.


We also compare the BLEU scores of the NNLM selection and the NMT selection.
To that end, we select top $20\%$ data and use it to fine-tune the noisy Paracrawl
baseline (P1), for the NNLM method and the proposed method, respectively.

We had to resolve an issue in the NNLM selection experiment. Recall that the trusted data we use
is from WMT newstest 2010-2011 and the development set we use for stopping the training
is WMT newstest 2012-2013. WMT newstests across years do not seem to be in the same domain,
as a result, the perplexity on devset never drops in training with trusted data.
This would be additional evidence that improvements from our proposed denoising approach is
unlikely from domain adaptation. In the end, we had to extract randomly 1000 lines out of the
 trusted data as the devset for early stopping and use the remaining as the trusted data
when training the denoised model $\widehat{\theta}$ that
is used to compute the noise scores (or data relevance in the NNLM case) by Eq~\ref{noise}.

The BLEU scores in Table~\ref{LM_vs_NMT} show the clear difference.
The NNLM method does not discern noise and thus the top selection would be as noisy
as the baseline data. As a result, fine-tuning the noisy baseline (P1) would not improve.
As a matter of fact, the patent BLEU drops over baseline, probably indicating that domain data selection
causes data bias. The proposed method, on the other hand, performs clearly better (P8),
for example, compared to P9, +2.5 BLEU on n2014, +3.8 BLEU on d2015 and +10.4 BLEU
on patent. These prove the effectiveness of the proposed method in producing better
systems on noisy data.


\begin{table}[t]
\centering
  \begin{tabular}{lccc}
    \hline
    System & n2014 & d2015& patent \\
    \hline
    \hline
    P1 Paracrawl baseline& 31.6 & 30.7 &  37.2 \\
    P8 P1+NMT $20\%$ & {\bf 34.3} & {\bf 34.7} & {\bf 45.8} \\
    P9 P1+NNLM $20\%$ & 31.8 & 30.5 & 35.4 \\
    \hline
  \end{tabular}
\caption{LM method does not denoise, but NMT method (proposed) does; and
         a denoised model has improved general translation accuracy.
         P1+NMT $20\%$: fine-tune P1 with top $20\%$ selection by NMT method.
         P1+NNLM $20\%$: fine-tune P1 with top $20\%$ selection by NNLM method.\label{LM_vs_NMT}}
\end{table}


\subsection{Discussion}
The research in \cite{dynamiccds} that selects data with
neural language models show that dynamically selected
parallel data for domain adaptation improves domain test sets, but it can hurt
test sets that are out of domain. It also shows that the dynamic online
selection still underperforms the fine-tuning on domain parallel data.
In our denoising results, the online denoising (e.g., P3) can significantly
outperform the simple fine-tuning (e.g., P2).

We clarify that our method could potentially work with
other data filtering methods. For example, if the underlying noisy data has
already been filtered, applying online denoising with trusted data could potentially
bring even further improvement than no pre-filtering.

\section{Conclusion and Future Research}\label{conclusions}
Domain data selection and domain adaptation for machine translation is a well-studied
topic, but denoising training data or MT training is not yet, especially for NMT training.
In this paper, we generalize the recipes of effective domain data research to
address a different and important data quality for NMT -- data noise. We define how to
measure noise and how to select noise-reduced data batches to train NMT models online.
We show that the noise we define correlates with human ratings and that the proposed
approach yields significantly better NMT models.


The method probably can be tried to denoising for other seq2seq tasks like parsing, image labeling.
It seems interesting to study and understand the properties that trusted data should have.
 It also sounds an interesting research to discover better data orders.






\section*{Acknowledgments}
The authors would like to thank George Foster for his help
refine the paper and advice on various technical isses in
the paper, Thorsten Brants for his earlier work on the topic,
 Christian Buck for his help with the Paracrawl data,
Yuan Cao for his valuable comments and suggestions on the paper,
and the anonymous reviewers for their constructive reviews.

\bibliography{emnlp2018}
\bibliographystyle{acl_natbib_nourl}

\appendix


\end{document}